%% file: main.tex
\documentclass[letterpaper, 10 pt, conference]{ieeeconf}

\IEEEoverridecommandlockouts

\input{macros.tex}

\title{\LARGE \bf  \ours{}: Distilling Vision–Language Reasoning into Attention Maps for Social Robot Navigation
\\}

\author{Mohamed Elnoor$^{1}$,  Kasun Weerakoon$^{1}$, Gershom Seneviratne$^{1}$, Jing Liang$^{2}$, Vignesh Rajagopal$^{3}$, \\ and Dinesh Manocha$^{1,2}$
\thanks{$^{1}$ Authors are with Dept. of Electrical and Computer Engineering, University of Maryland, College Park, MD, USA. {\tt\footnotesize melnoor@umd.edu, 
    kasunw@umd.edu, gershom@umd.edu.} }
    \thanks{$^{2}$ Authors are with Dept. of Computer Science, University of Maryland, College Park, MD, USA. {\tt\footnotesize jingl@umd.edu, dmanocha@umd.edu} }
    \thanks{$^{3}$ Author is with James Clark School of Engineering, University of Maryland, College Park, MD, USA. {\tt\footnotesize vigneshr@umd.eduu} }
}

\begin{document}

\maketitle

\begin{abstract}



We introduce \ours{}, a novel method for distilling vision–language reasoning from large Vision–Language Models (VLMs) into spatial attention maps for socially compliant robot navigation. Unlike traditional methods that rely on expert demonstrations or human-annotated datasets, \ours{} performs knowledge distillation and fine-tuning at the intermediate layer representation (attention) level by aligning attention maps from a pretrained vision-action model with socially guided attention maps derived from a large VLM. These distilled attention maps highlight key navigational regions in a scene and serve as socially informed spatial cost maps for motion planning. To achieve this, we introduce a novel attention-level distillation loss that fuses knowledge from both sources, generating augmented attention maps with enhanced social awareness. These refined attention maps are then used as a traversability costmap within a socially aware local planner for navigation. We validate our approach through real-world experiments on a Husky wheeled robot, and demonstrate 14.2\% - 50\% improvements in success rate over existing methods. 


\end{abstract}


\input{1_Introduction}

\input{2_Related_Works}
\input{3_Background}

\input{4_approach}
\input{6_Results}

\input{7_Conclusion}

\bibliographystyle{IEEEtran}
\bibliography{references}

\end{document}

%% file: macros.tex
\usepackage{xcolor}
\usepackage{amssymb}
\usepackage{graphicx}
\usepackage{mathtools}
\usepackage{amsmath}
\usepackage{gensymb}
\usepackage{float}
\usepackage{multirow}
\usepackage{makecell}
\usepackage{mathtools}
\usepackage{hyperref}
\usepackage{overpic}
\usepackage{subcaption}
\usepackage{xcolor}
\hypersetup{
    colorlinks=true,
    linkcolor=blue,
    filecolor=magenta,      
    urlcolor=blue,
    pdftitle={Overleaf Example},
    pdfpagemode=FullScreen,
    }

\usepackage{tabularx}
    \newcolumntype{L}{>{\raggedright\arraybackslash}X}
\usepackage[utf8]{inputenc}
\usepackage[english]{babel}

\usepackage{amsthm}
\usepackage{booktabs}
\usepackage{algorithm}
\usepackage{algpseudocode}

\usepackage{subcaption}

\newcommand{\ours}{ViLAM}

\newcommand{\no}{\noindent}



%% file: 1_Introduction.tex

\section{Introduction}  \label{sec:Intro}

As autonomous robots become increasingly integrated into human-centered environments, ensuring safe, efficient, and socially compliant navigation is a critical challenge \cite{francis2023principles}. Robots are deployed in a wide range of real-world applications, including service robotics, delivery, hospital logistics, office maintenance, elderly care, and urban mobility, where they must operate alongside pedestrians and navigate complex, dynamic environments \cite{holland2021service, chen2021adoption, nagatani2021innovative, bardaro2022robots}. Unlike structured industrial settings, these environments are inherently unpredictable and require robots to interpret social cues, anticipate human movement, and react appropriately in real time. 

However, traditional navigation methods primarily focus on collision avoidance and geometric path planning \cite{daza2021approach}, often treating humans as static or moving obstacles rather than interactive agents with social expectations \cite{singh2022understanding}. This limitation can lead to unnatural and sometimes disruptive robot behaviors, such as cutting through groups of people, blocking pathways, or failing to yield in shared spaces \cite{mavrogiannis2023core}.

To address these challenges, researchers have explored socially aware navigation, where robots must understand and respond to human behaviors and environmental context \cite{gao2022evaluation}. Prior approaches rely on hand-crafted social rules \cite{singamaneni2024survey}, imitation learning from human demonstrations \cite{qin2021deep}, or reinforcement learning-based policies \cite{samsani2021socially}. While these methods have demonstrated promising results, they often require large-scale training data and struggle to generalize to unseen environments. Moreover, many approaches lack high-level semantic reasoning, limiting a robot’s ability to infer context-appropriate behaviors in complex social settings.

\begin{figure}[t]
      \centering  \includegraphics[width=0.9\columnwidth,]{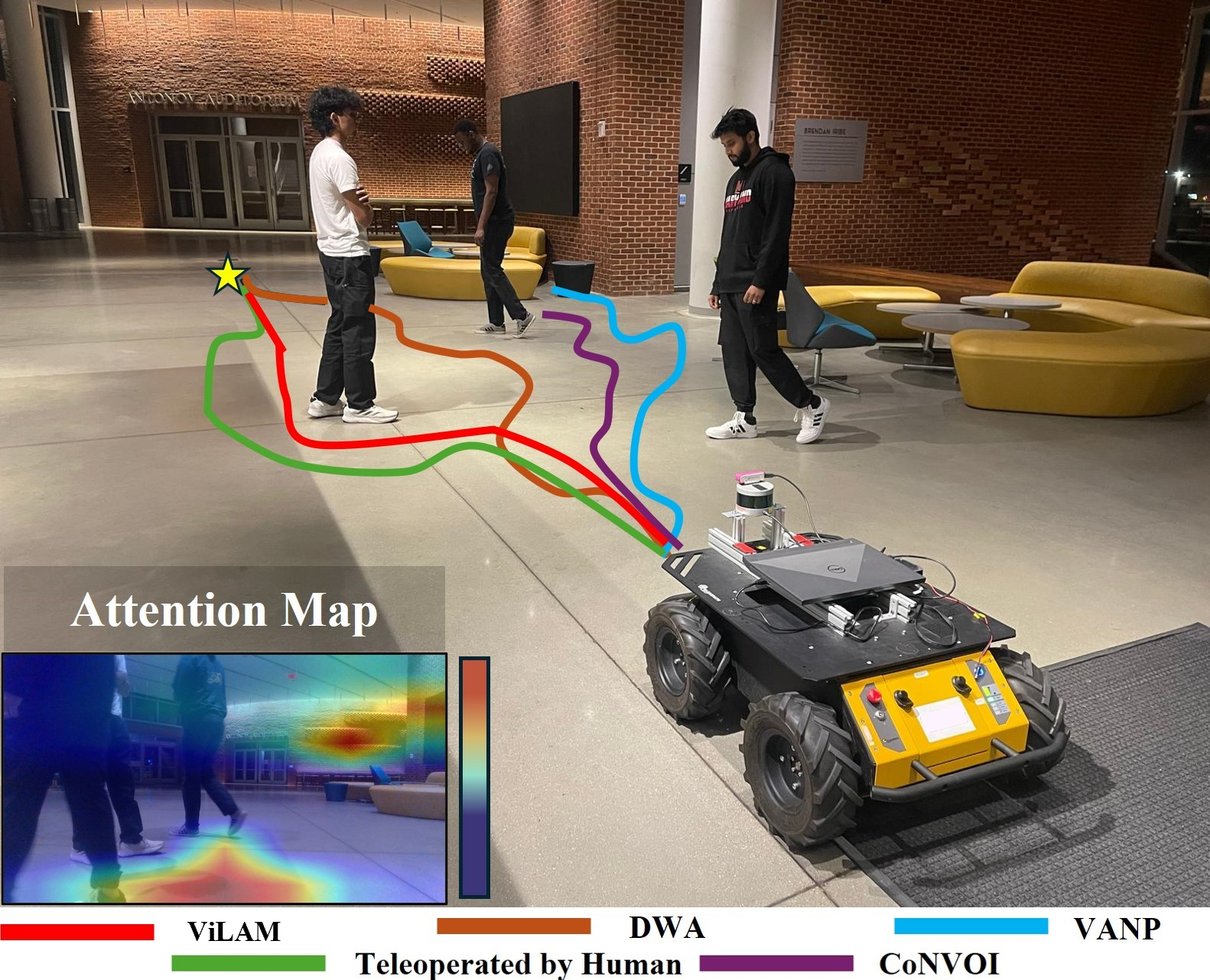}
      \caption {\small{ Robot navigation using \ours{} and baseline methods in a social navigation scenario. \ours{} distills social navigation knowledge from a pretrained vision-action model VANP \cite{10802451} and a large VLM, by leveraging attention maps to generate an enhanced attention representation for socially compliant navigation. This improved attention representation (\textbf{LEFT BOTTOM}) enables better understanding of human intentions, allowing the robot to anticipate movement patterns and avoid potential disruptions to pedestrians.}}
      \label{fig:cover}
      \vspace{-20pt}
\end{figure}

Recent advances in Vision-Language Models (VLMs) and Large Language Models (LLMs), such as GPT-4V \cite{openai2024gpt4technicalreport}, Gemini \cite{geminiteam2024geminifamilyhighlycapable}, and LLaMA \cite{touvron2023llama}, have demonstrated remarkable capabilities in semantic scene understanding, commonsense reasoning, and human behavior prediction \cite{wu2024vision, 10160969, weerakoon2024behav}. These models exhibit strong zero-shot and few-shot reasoning abilities that extend beyond conventional perception tasks such as object detection or segmentation. In navigation scenarios, VLMs can infer implicit cues about human motion and social context, such as identifying regions pedestrians are likely to occupy or areas that should be avoided to maintain social comfort \cite{gorlo2024long, payandeh2024social, 10802716}. 

However, the large model size of modern VLMs—often consisting of billions of parameters—creates significant challenges for real-time deployment on resource-constrained robotic platforms. Running VLM inference onboard a mobile robot requires substantial memory and computational resources, which can introduce latency during decision-making. Since navigation in dynamic environments requires rapid responses to avoid collisions and maintain social compliance, directly executing VLMs during robot navigation is often impractical for real-time safety-critical systems.


\textbf{Main contributions:} To address these challenges, we introduce \ours{}, a method for distilling vision--language reasoning from large Vision-Language Models (VLMs) into spatial attention maps for socially aware robot navigation. Instead of relying on continuous queries to a large VLM during deployment, \ours{} distills socially relevant navigation cues into a lightweight transformer-based model.  Specifically, our method aligns attention maps from a pretrained vision-action model with socially guided attention maps generated by a VLM. The resulting distilled attention maps are then used as spatial cost maps for real-time motion planning. This enables efficient and socially compliant navigation while avoiding the computational overhead of online VLM inference. Our key contributions are:

\begin{itemize}
    \item  \textbf{Distilling Vision-Language Reasoning into Attention Maps:} We propose a novel attention-based distillation method that transfers socially compliant navigation reasoning from a large Vision-Language Model (VLM) and a pretrained vision-action model \cite{10802451} into a lightweight transformer-based model. Unlike traditional knowledge distillation, which focuses on output predictions, our approach distills knowledge at the attention map level by aligning intermediate attention representations from the vision-action model with attention-like semantic maps derived from the VLM. While \cite{10802451}  produces attention-like activations as an emergent property of self-supervised pretraining, our method grounds those activations in VLM semantics through knowledge distillation. This joint attention distillation enables \ours{} to learn enhanced attention representations, leveraging knowledge from both models. As a result, \ours{} produces trajectories that are 28.7\% closer to human teleoperated actions in terms of Fréchet distance, ensuring improved social compliance and alignment with human navigation behavior. 
    
    %

    \item \textbf{Socially-Guided Attention Fine-Tuning:} We introduce a Structural Similarity Index (SSIM) Loss to align \ours{}'s attention map predictions with both the pretrained vision-action model's attention maps and the attention-like semantic maps from the large VLM. Our cosine similarity-based SSIM loss formulation ensures smoother gradient updates, leading to more stable and effective learning of socially relevant attention regions. This results in a 14.2\% - 50\% improvement in navigation success rate.
    

    \item \textbf{An Adapted Local Motion Planner for Smooth and Social Navigation:} To translate \ours{}'s attention cost map into real-time motion control, we integrate a Dynamic Window Approach (DWA) \cite{dwa} planner that dynamically adjusts robot movement based on the distilled attention maps. This planner ensures that the robot’s navigation adheres to social norms while maintaining smooth, natural, and real-time trajectories in real-world environments.

    \end{itemize}

Through extensive evaluations, we demonstrate that \ours{} enables robots to navigate efficiently while adhering to social norms, achieving higher success rates and smoother motion execution compared to existing baselines. 

%% file: 2_Related_Works.tex

\section{Related Work}

In this section, we review existing approaches for robot navigation in dynamic environments. We also discuss the role of large pretrained models in navigation and knowledge distillation techniques for efficient deployment in resource-constrained systems.

\subsection{Robot Navigation in Dynamic Scenes}
Navigating dynamic environments requires robots to balance obstacle avoidance, goal achievement, and social compliance. Both classical and learning-based approaches have been developed to address these challenges, each with distinct advantages and limitations.

\subsubsection{Classical Navigation Methods}
Classical navigation methods rely on optimization-driven and rule-based strategies. Model Predictive Control (MPC) and its derivatives predict future system behaviors by solving optimization problems over a finite horizon to ensure collision-free and smooth trajectories \cite{sugimoto2024mobile, brooks2009randomised, lafmejani2021nonlinear,fox1997dynamic}. Furthermore, Velocity Obstacle (VO)-based methods predict potential collisions by modeling the movement of surrounding agents, which enable robots to adjust their paths accordingly \cite{zhang2017dynamic}. 
However, these methods often lack adaptability to complex human behaviors as they depend on predefined models and constraints. Although they excel at static obstacle avoidance, they fail to incorporate social norms, which leads to unnatural robot behavior in human-centered environments.

\subsubsection{Learning-Based Navigation Methods}
To address the limitations of classical approaches, learning-based methods leverage data-driven models to improve adaptability. Unlike rule-based methods, these models learn navigation behaviors directly from the data, allowing robots to generalize to more diverse environments \cite{mavrogiannis2023core}. Imitation Learning (IL) enables robots to mimic expert demonstrations, facilitating the acquisition of human-like movement patterns \cite{yildirim2022learning, torabi2018behavioral, yan2022mapless}. Reinforcement Learning (RL) techniques
allow robots to learn optimal navigation policies through continuous interaction with the environment \cite{zhu2021deep}. In addition, inverse reinforcement learning (IRL) has been used to infer human navigation preferences by learning cost functions from observed trajectories \cite{kim2016socially, tung2018socially, perez2021robot}. Graph Neural Networks (GNNs) and transformer-based policies have also been explored to encode long-range spatial relationships for efficient decision-making in dynamic spaces \cite{8729387}.
However, these methods face challenges in real-world applications. IL models are limited by the quality and diversity of their training data. 
RL and IRL models require extensive training and may not work well due to domain transfer issues from simulated to real environments \cite{zhao2020sim}. 

\begin{figure*}[t]
      \centering  \includegraphics[width=0.85\linewidth]{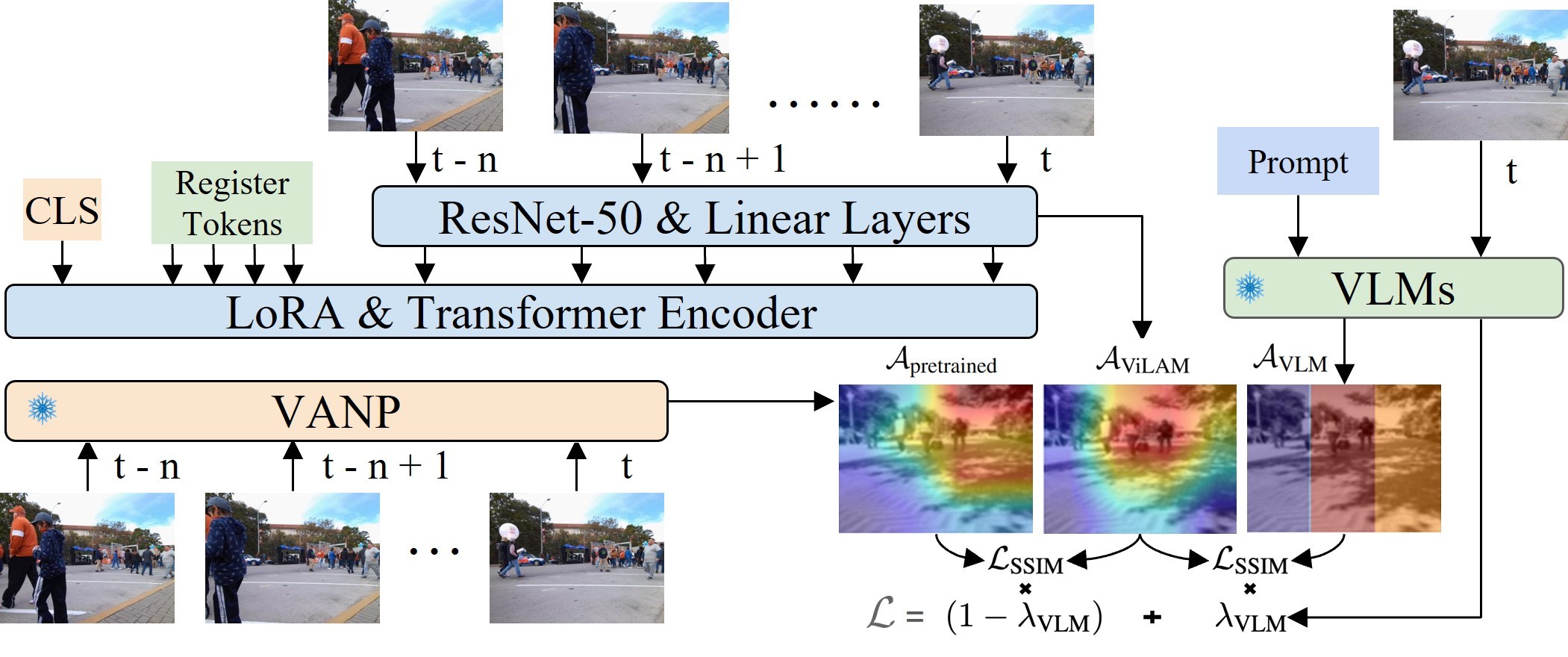}
      \caption {\small{ System architecture of \ours{}. Our method distills social navigation knowledge from a pretrained vision-action model (VANP) \cite{10802451} and a large Vision-Language Model (VLM) by aligning their attention maps, rather than performing end-to-end distillation or fine-tuning. These attention maps highlight critical regions for socially compliant navigation and are extracted from intermediate layers of the image encoders. We adopt the same CNN backbone (ResNet-50) and Transformer encoder as VANP. Classification (CLS) and learned register tokens are appended to image embeddings to encode global scene context and enrich spatial attention reasoning. \ours{} employs Structural Similarity Index (SSIM) loss to effectively distill attention information from both VANP’s intermediate attention layers and the predictive attention maps of a large VLM. }
      }
      \label{fig:system-arch}
      \vspace{-15pt}
\end{figure*}

\subsection{Large Models for Navigation}

Recent advances in large pretrained models (LPMs), including large language models (LLMs) and vision-language models (VLMs), have improved robotic perception, planning, and navigation \cite{luddecke2022image, kuo2022f, elnoor2024robot}.
VLMs have been widely adopted to improve robotic perception, particularly in semantic segmentation and object recognition. CLIPSeg \cite{luddecke2022image} enables zero-shot segmentation, allowing robots to identify objects and landmarks based on text prompts, which supports goal-conditioned navigation. VLMaps \cite{10160969} integrates vision language features into 3D spatial maps, providing natural language-guided navigation and improving scene understanding for better localization and interaction with the environment.

In addition, VLMs have been applied to navigation tasks across diverse environments \cite{zeng2023large}. For instance, LM-Nav \cite{shah2023lm} combines GPT-3 and CLIP to interpret natural language commands and visual cues, enabling the generation of efficient navigation paths. This integration of vision-language reasoning supports high-level robotic planning and real-world deployment.  ViNT \cite{shah2023vint} introduces a transformer-based foundation model trained on diverse navigation datasets, which allowed the robot to generalize across different environments and tasks, such as long-horizon planning and human-aware pathfinding. Similarly, L3MVN \cite{10342512} leverages large language models (LLMs) for visual target navigation, that uses language-based frontier selection to improve efficiency and generalization in unknown environments.
In dynamic human-centered spaces, VLM-Social-Nav \cite{10777573} employs a GPT-based scoring module to compute a social cost function, which guides robots to socially appropriate actions. Moreover, CoNVOI \cite{10802716} leverages visual annotations to extract waypoint sequences from camera observations, ensuring that robots follow context-aware navigation behaviors in dynamic environments.
Despite their promising capabilities in understanding social norms, predicting human motion, and adapting to dynamic environments, VLMs impose significant computational demands, restricting real-time deployment on resource-constrained robots using edge hardware. Additionally, inference times vary depending on network dependencies, particularly when relying on cloud-based processing, which can reduce responsiveness in robot navigation in dynamic and complex environments. Our approach addresses these challenges by distilling social reasoning from Vision-Language Models (VLMs) into a lightweight, deployable model.

\subsection{Knowledge Distillation}

Knowledge distillation transfers knowledge from a large, computationally intensive teacher model to a smaller, efficient student model, a process that facilitates real-time inference on resource-constrained systems \cite{goh2023self}. In robot navigation, compact models acquire high-level reasoning capabilities from extensive models \cite{sanh2019distilbert}, capabilities such as semantic scene understanding and human behavior prediction, without incurring significant computational costs \cite{tosi2020distilled}.
Recent studies apply knowledge distillation in robotics using various methodologies. One approach learns human-like collision avoidance policies in decentralized multi-agent environments by transferring knowledge from human demonstrations to agent models \cite{Xu_2021}. Another approach transfers first-person-view representations into universally applicable third-person-view representations, a strategy that improves navigation models' adaptability across different platforms \cite{uemura2024lmdpgncrossmodalknowledgedistillation}.
Nevertheless, existing techniques often overlook the integration of social reasoning into navigation policies. Large Vision-Language Models (VLMs), such as GPT-4V \cite{openai2024gpt4technicalreport} and LLaVA \cite{liu2023llava}, provide good capabilities in terms of human behavior prediction; however, their substantial computational demands and latency hinder real-time applications. Meanwhile, Attention maps are widely studied as a means of interpreting transformer models  \cite{clark2019does, dosovitskiy2020image, chefer2021transformer}. In vision tasks, they highlight salient regions that align with human perception \cite{dosovitskiy2020image}. In navigation, Nazeri et al. \cite{10802451} found that mid-level attention activations from pretrained models highlight navigationally relevant areas, even without explicit supervision. Building on these insights, our work performs attention-level distillation from both vision-language and pretrained vision-action models.


%% file: 3_Background.tex
\section{Background}

In this section, we explain our approach, state our assumptions, and introduce
key concepts used in our work.

\subsection{Setup and Conventions}
Our formulation assumes a ground robot equipped with an RGB camera with a common coordinate frame centered at the robot’s center of mass. The $X$, $Y$, and $Z$ axes point forward, left, and up, respectively. The camera provides RGB images $I_{\text{RGB}, t}$. Additionally, the IMU provides orientation and motion feedback.
The robot uses a controller architecture that receives linear and angular velocity commands $(v, \omega)$ in the robot’s frame. 
\subsection{Pretrained Model} \label{sec: pretrained}

Pretrained models aid robot navigation by extracting visual features, but conventional models like ImageNet-trained ones focus on object detection rather than navigation-specific cues like paths, obstacles, and human movement. While self-supervised learning (SSL) addresses this, many approaches require large datasets or lack generalization. We adopt VANP \cite{10802451}, which is a self-supervised vision-action model trained to predict future actions from past observations. Instead of generic object-centric representations, VANP extracts navigation-relevant features by aligning visual observations with action trajectories. The model processes a sequence of \( n \) past images and generates an attention map \( \mathcal{A}_{\text{pretrained}} \), which highlights regions critical for navigation. This can be expressed as:

\begin{equation}
    \mathcal{A}_{\text{pretrained}} = \mathcal{F}_{\text{pretrained}}(I_{\text{RGB}, t-n}, \dots, I_{\text{RGB}, t})
\end{equation}

where \( \mathcal{F}_{\text{pretrained}} \) represents the pretrained model, and \( I_{\text{RGB}, t-n}, \dots, I_{\text{RGB}, t} \) denote the sequence of past RGB images used as input.

%% file: 4_approach.tex
\section{Our Approach}


In this section, we present \ours{}, a method for socially aware robot navigation. \ours{} enables real-time, resource-efficient navigation by distilling social reasoning from large Vision-Language Models (VLMs) into a compact transformer-based model. Our method consists of four key components:  

\begin{itemize}  
    \item \textbf{Data Generation}: A dataset is constructed by leveraging VLMs to generate socially guided attention maps that highlight navigation-relevant regions.  
    \item \textbf{Distilled Model}: A lightweight transformer-based model is fine-tuned to align with VLM-derived attention maps to ensure socially compliant decision-making.  
    \item \textbf{Attention-Guided Loss Function}: An Attention consistency loss function refines model attention to balance trajectory learning with social awareness.  
    \item \textbf{Socially Aware Motion Planner}: A Dynamic Window Approach (DWA) planner integrates the distilled attention to generate smooth and adaptive robot movement.  
\end{itemize}  

The overall system architecture is shown in Fig. \ref{fig:system-arch}. The following subsections describe each component in detail.

\subsection{Data Generation} \label{sec: Data_Generation}

Our approach constructs a socially guided navigation dataset using VLM-based supervision. We select a customized subset of SCAND, a large-scale socially compliant navigation dataset \cite{karnan2022socially}, and annotate images with VLM reasoning.
To generate the customized dataset, we leverage a frontier-based evaluation that allows the distilled model to infer socially aware navigation cues without direct VLM queries. For each sample, we define three navigation frontiers (left, center, and right) within the RGB image $I_{\text{RGB}, t}$. These frontiers represent possible regions the robot could take while accounting for pedestrian movement and social norms. The marking process overlays colored rectangles on each frontier to ensure the VLM evaluates them explicitly.

We query a large VLM with a Chain-of-Thought (CoT) prompting approach \cite{wei2022chain}. Given the image, social context, and visible pedestrians, the VLM estimates the likelihood of each frontier becoming crowded. The likelihood score follows the equation:

\begin{equation}
P(f) = \text{VLM}(I_{\text{RGB}}, \mathcal{T}_{\text{prompt}}),
\end{equation}

 \noindent where $f \in \{\text{left, center, right}\}$ represents the navigation frontiers, and $\mathcal{T}_{\text{prompt}}$ is a structured query designed to elicit social context reasoning.

 Based on the likelihood estimates, we annotate $I_{\text{RGB}}$ to produce $\mathcal{A}_{\text{VLM}}$, encoding socially guided navigation cues. The dataset follows the structure:

\begin{equation}
\mathcal{D}_{\text{VLM}} = \{ (I_{\text{RGB}}^{m}, \mathcal{A}_{\text{VLM}}^{m}) \mid m = 1, \dots, 10k\},
\end{equation}

 \noindent where \( D_{\text{VLM}} \) represents the VLM-annotated image dataset that incorporates human behavior, movement patterns, and social norms. 
By using VLM reasoning, we generate socially aware annotations offline, which avoids expensive computation during inference. This dataset trains the distilled model to capture social navigation behaviors while staying efficient for deployment.

\subsection{Distilled Model}

To efficiently adapt a pretrained vision model for socially aware navigation, we employ Low-Rank Adaptation (LoRA), a parameter-efficient fine-tuning approach. Instead of updating all model parameters, LoRA introduces low-rank trainable adapters while keeping the original model weights frozen. This significantly reduces computational overhead and memory usage while maintaining the pretrained model’s expressivity. As shown in figure \ref{fig:system-arch}, our architecture consists of three parallel pipelines during training:

\begin{itemize}
    \item Pretrained Model (\(\mathcal{F}_{\text{pretrained}}\)): The frozen vision model extracts navigation-relevant attention maps, denoted as \(\mathcal{A}_{\text{pretrained}}\), as described in section \ref{sec: pretrained}.
    \item Distilled Model (\(\mathcal{F}_{\text{\ours{}}}\)): A copy of the pretrained model augmented with LoRA adapters fine-tunes a set of low-rank parameters. We extract the updated attention map (\(\mathcal{F}_{\text{\ours{}}}\))\  from the last layer of the ResNet50. 
    \item VLM Supervision : A large Vision-Language Model (VLM) processes the RGB image to generate socially guided attention maps, \(\mathcal{A}_{\text{VLM}}\), providing a supervisory signal for fine-tuning.
\end{itemize}

Our model optimizes (\(\mathcal{F}_{\text{\ours{}}}\)) parameters to balance two objectives: preserving the pretrained model’s navigation-aware representations while aligning with socially guided attention from the VLM.

\subsection{Attention-Guided Loss Function} 

To integrate social reasoning from Vision-Language Models (VLMs) while maintaining the navigation priors of the pretrained model, we introduce an attention consistency loss function. This loss encourages the fine-tuned model to retain critical navigation-relevant features while incorporating socially guided attention cues.

Given an input image \(I_{\text{RGB}}\), we extract the attention map from the frozen pretrained model, denoted as \(\mathcal{A}_{\text{pretrained}}\). Simultaneously, the fine-tuned model generates an updated attention map, \(\mathcal{A}_{\text{\ours{}}}\), which we seek to optimize. The socially guided attention map provided by the VLM is represented as \(A_{\text{VLM}}\). The total loss function is formulated as:
\begin{align}
\mathcal{L} &= (1 - \lambda_{\text{VLM}}) \cdot \mathcal{L}_{\text{SSIM}}(\mathcal{A}_{\text{\ours{}}}, \mathcal{A}_{\text{pretrained}}) \nonumber \\
&\quad + \lambda_{\text{VLM}} \cdot \mathcal{L}_{\text{SSIM}}(\mathcal{A}_{\text{\ours{}}}, \mathcal{A}_{\text{VLM}}),
\end{align}

\no where the $ \mathcal{L}_{\text{SSIM}}(A, B)$ is defined as Cosine Similarity between flattened attention maps $A$ and $B$. 



This facilitates the integration of human-aware navigation patterns while maintaining consistency with the pretrained trajectory-based attention. 

\subsection{Socially Aware Motion Planner} \label{sec: planner}
We adopt a modified local planner, inspired by \cite{fox1997dynamic}, to generate socially aware navigation trajectories. Our planner optimizes linear and angular velocity pairs \((v, \omega)\) that guide the robot toward its goal while ensuring smooth and socially compliant movement. Our method integrates a novel social cost function to align the robot’s behavior with human navigation norms by leveraging the Attention map ($\mathcal{A}_{\text{\ours{}}}$).

The planner first determines an admissible velocity space \(\mathcal{V}_s\). From which we derive a reduced set of feasible velocities \(\mathcal{V}_r\) by filtering out dynamically infeasible or collision-prone trajectories.
To compute the optimal control action from \(\mathcal{V}_r\), we define an objective function:

\begin{equation}
    J(v,\omega) = \beta_1 \cdot goal(v,\omega) + \beta_2 \cdot soc(v,\omega)
\end{equation}

 \noindent where:

\begin{itemize}
    \item \( goal(v,\omega) \) encourages movement toward the target.
    \item \( soc(v,\omega) \) introduces a social cost function that depends on the Attention map ($\mathcal{A}_{\text{\ours{}}}$).
\end{itemize}

We define the social cost function \( soc(v,\omega) \) by computing the alignment between the projected trajectory and the distilled attention map \( \mathcal{A}_\text{\ours{}} \). Given an action \((v, \omega)\), we first extrapolate a short-horizon trajectory \(traj^R(v, \omega)\) relative to the robot’s frame using the unicycle kinematic model, where the state evolves according to \(\dot{x} = v \cos(\theta)\), \(\dot{y} = v \sin(\theta)\), and \(\dot{\theta} = \omega\). This trajectory is then projected onto the attention cost map  \(\mathcal{A}_\text{\ours{}}\) using homography projection to compute the social cost.

\begin{equation}
        soc(v, \omega) = \max_{(i,j)\in traj^C(v,\omega)} \mathcal{A}_\text{\ours{}}(i,j),
\end{equation}

\noindent where \( traj^C(v,\omega) \) is the trajectory projected onto the cost map, and \( \mathcal{A}_\text{\ours{}}(i,j) \) represents the distilled attention cost map at a given location.

Finally, the optimal control action \((v^*,\omega^*)\) is obtained by minimizing the objective function:

\begin{equation}
    (v^*,\omega^*) = \underset{(v,\omega) \in \mathcal{V}_r}{\operatorname{argmin}} J(v,\omega)
\end{equation}


\begin{figure}[t]
      \centering  \includegraphics[width=\columnwidth]{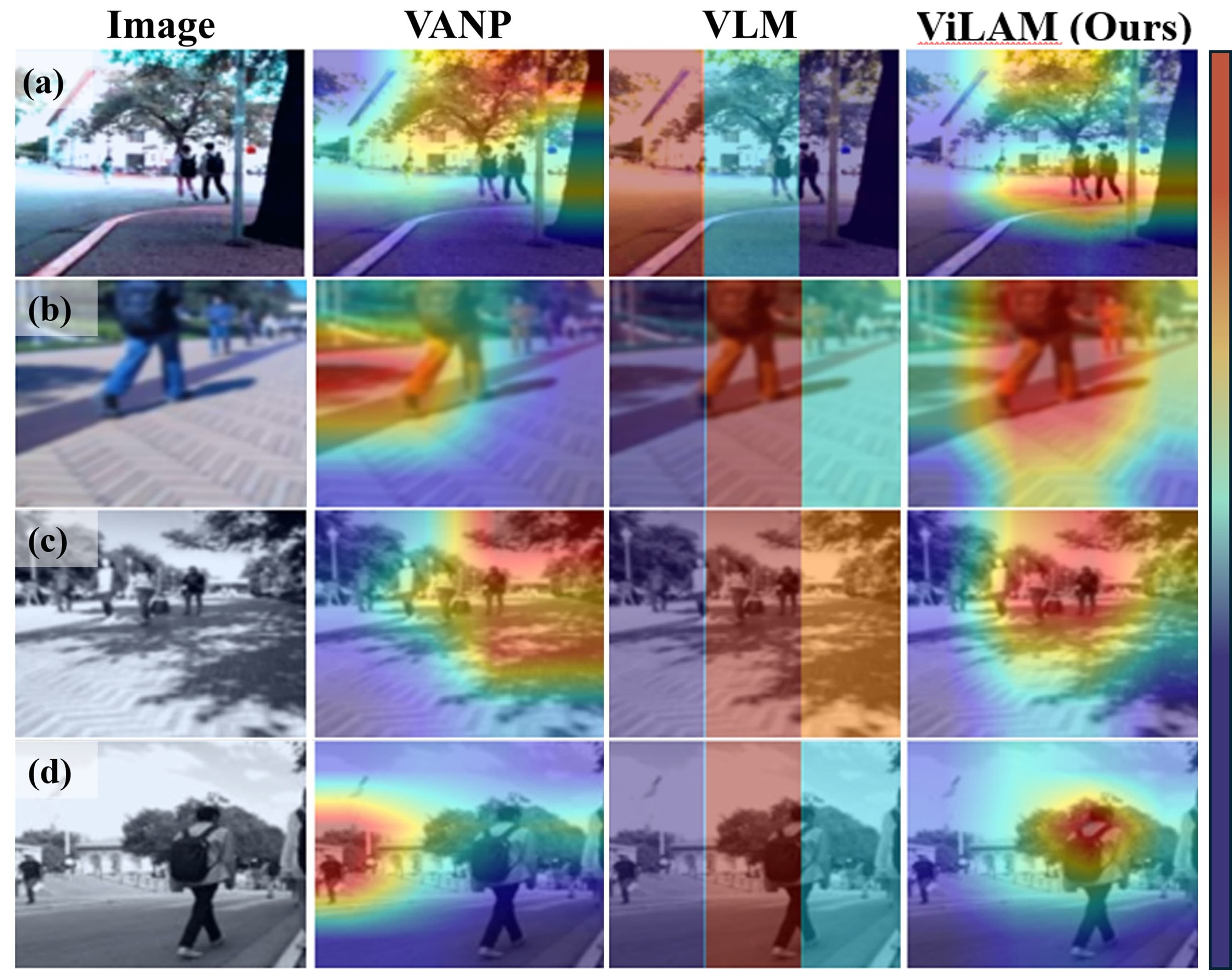}
      \caption {\small{ Attention maps generated using our method \ours{} by distilling attention knowledge from both pretrained vision-action model VANP \cite{10802451}, and the large VLM in different social scenarios compared to the attention maps from VANP and the large VLM. These attention maps are compared against those from VANP and the large VLM. The Jet color map is applied to highlight attended regions, with red indicating the most highly attended areas. \ours{} demonstrates improved attention over both the pretrained model and the large VLM. By leveraging combined knowledge through attention distillation, \ours{} effectively corrects missed attention from both sources. This leads to enhanced focus on critical objects and regions within a scene. 
       }}
      \label{fig:heatmap_samples}
      \vspace{-20pt}
\end{figure}

%% file: 6_Results.tex
\section{Results and Analysis}

\begin{figure*}[t]
      \centering  \includegraphics[width=2\columnwidth, height = 4cm]{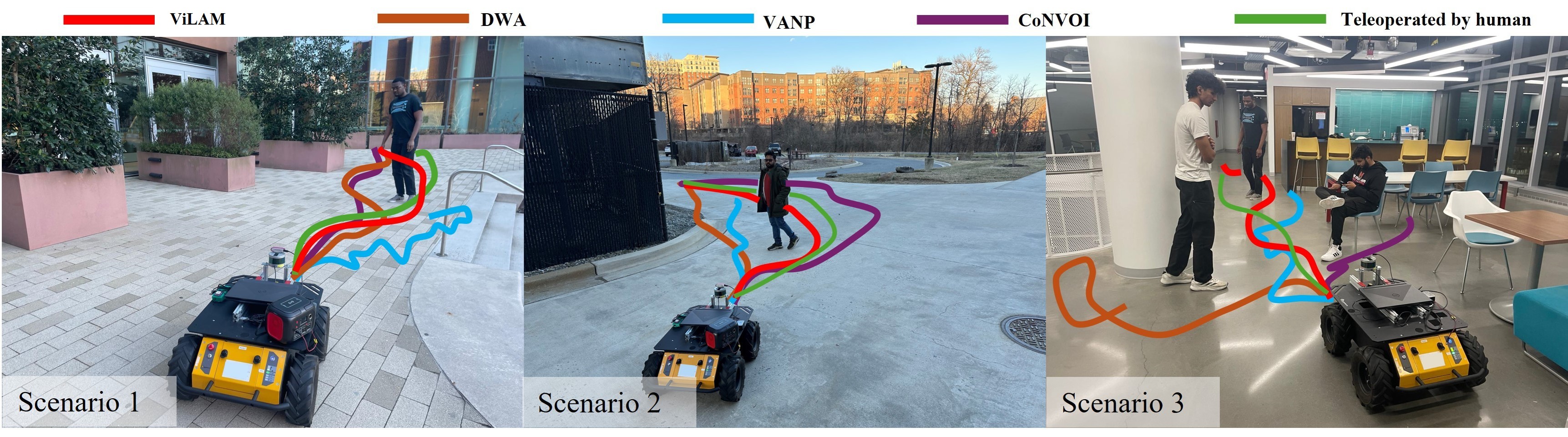}
      \caption {\small{
        Robot trajectories in complex social navigation scenarios where the directional intent of agents needs to be taken into account during planning. Our method identifies dynamic motion and navigational intent of agents within the environment based on the distilled attention maps, which are used to plan in an improved socially compliant manner without disrupting agent motion within the environment.  For example, in scenario 1, DWA and CoNVOI fail to anticipate motion, while in scenario 2, VANP and DWA exhibit the same limitation.       
      } 
      }
      \label{fig:scn-trajs}
      \vspace{-20pt}
\end{figure*}

\subsection{Implementation \& Robot Setup}

We implement our method using PyTorch and train the model on an Nvidia A6000 GPU. 
For real-world experiments, we use a Clearpath Husky robot equipped with an Intel RealSense D435i camera, and an onboard laptop with an Intel i9 processor and an Nvidia RTX 2080 GPU. We use GPT-4o to annotate the dataset described in section \ref{sec: Data_Generation}.

\subsection{Comparison Methods}
We compare our method with classical, learning-based and VLM-based methods.
\begin{itemize}

    \item \textbf{Dynamic Window Approach (DWA)} \cite{fox1997dynamic}: A single step MPC-based method that uses 2D LiDAR scans for obstacle avoidance.
        

    \item \textbf{CoNVOI} \cite{10802716}: A context-aware VLM-based navigation method using vision-language models to generate reference trajectories, which is executed by a classical planner.
    
    \item \textbf{VANP} \cite{10802451}: A self-supervised pretrained model that's used to generate the attention map ($\mathcal{A}_{\text{pretrained}}$). We integrate this attention map with our planner described in section \ref{sec: planner}.


\end{itemize}


\subsection{Evaluation Metrics}\label{sec:metrics}

\begin{itemize}
    \item \textbf{Success Rate}: The ratio of successful navigation trials where the robot was able to reach its goal without freezing or colliding with obstacles.


    \item \textbf{Time to Goal}: The robot’s average time (in seconds) to reach its goal in the successful trials.

    \item \textbf{Fréchet Distance}: Measures the Fréchet distance \cite{alt1995frechet_distance} (a measure of similarity between two curves) between a human teleoperated robot trajectory versus a comparison method's trajectory.

\end{itemize}

\subsection{Test Scenarios}

 We note that we use only the outdoor trajectories to train our distilled model in section \ref{sec: Data_Generation}. However, we compare all the methods in both outdoor and indoor scenarios:
\begin{itemize}
    \item \textbf{Scenario 1:} contains a dynamic human agent and static obstacles. 
    \item \textbf{Scenario 2:} contains a dynamic human agent, fence, low curb.
    \item \textbf{Scenario 3:} multiple human agents (sitting and walking), a nested fence pillar, chairs and tables.
    \item \textbf{Scenario 4:} multiple human agents walking under different lighting conditions.
\end{itemize}

\subsection{Analysis and Discussion}
We evaluated our method’s navigation performance qualitatively in Figs. \ref{fig:cover}, \ref{fig:scn-trajs} and quantitatively in Table \ref{tab:tab2}. We conduct experiments across four different real-world scenarios, two outdoor and two indoor. The outdoor scenarios involve dynamic and static obstacles, such as pedestrians and low curbs, while the indoor scenarios feature human presence in low-light conditions. Across all scenarios, \ours{} consistently achieves the highest success rate in goal-reaching by effectively navigating around obstacles. Additionally, our method exhibits behavior closely resembling human teleoperation, as evidenced by the lower Fréchet distance across all scenarios compared to other navigation methods.

 \no \textbf{In Scenario 1}: This scenario involves a dynamic agent and static obstacles. We observe that \ours{} initiates movement earlier compared to DWA, which tends to approach the human more closely before adjusting its trajectory. Additionally, CoNVOI takes a longer time to reach the goal due to the inference latency caused by querying the VLM over the internet. In contrast, \ours{} efficiently reaches the goal while maintaining a human-like trajectory, as evidenced by its low Fréchet distance. Furthermore, our method accurately predicts dynamic agent motion, which prevents unnecessary detours or abrupt trajectory shifts, unlike DWA and CoNVOI.

 \no \textbf{In Scenario 2}: This scenario includes a short curb, which poses a challenge for LiDAR-only methods like DWA, which led to frequent collisions and a lower success rate. CoNVOI also struggles in certain trials due to its reliance on LiDAR for planning, occasionally failing to detect the curb. In comparison, \ours{} accurately identifies the curb and adjusts its trajectory by turning right, effectively avoiding both the obstacle and the human. Our method demonstrates enhanced scene understanding and adaptive planning by following a trajectory closely resembling that of a human operator.

\no \textbf{In Scenario 3}: This scenario presents various obstacles, including chairs, tables, a nested fence, and both stationary and moving humans. VANP and DWA struggle to navigate effectively, frequently colliding with the nested fence and walking individuals. In contrast, our method successfully maneuvers through the obstacles while maintaining awareness of and adapting to human presence in the environment. \ours{} dynamically updates its attention map, highlighting potential regions of interest with respect to human movement, which enabled more socially compliant navigation compared to other baselines.

\no \textbf{In Scenario 4}: This scenario introduces challenging lighting variations that significantly impact vision-based navigation. VANP's performance deteriorates due to its reduced perception accuracy, which led to multiple collisions. Similarly, CoNVOI experiences difficulty under changing lighting conditions, requiring frequent VLM queries that prolong the navigation time. Although DWA avoids collisions, it exhibits a high Fréchet distance, often navigating uncomfortably close to humans. In contrast, our method dynamically adjusts its trajectory in response to moving obstacles, ensuring both safe and efficient navigation.

\no \textbf{Benefits of VLM}:
We note that leveraging VLM annotations played a crucial role in shaping the final attention map $\mathcal{A}_{\text{\ours{}}}$. By integrating VLM reasoning during training (offline), our method effectively utilizes its scene understanding capabilities to highlight relevant regions of interest based on contextual cues. After training, our method no longer queries the VLM during inference, where it relies solely on the distilled attention maps. This enhances interpretability and enables a structured, context-aware navigation method. Compared to using only the pretrained model (VANP) attention map $\mathcal{A}_{\text{pretrained}}$, our approach retains richer semantic understanding, leading to more informed decision-making and improved navigation performance across diverse scenarios.

\no \textbf{Benefits of sequential images}: 
We evaluate the impact of removing previous image sequences during training and observe that it negatively affects navigation performance across all scenarios, with particularly notable degradation in Scenarios 3 and 4. These scenarios involve highly dynamic human movement and significant lighting variations, where the absence of temporal context limits the model's ability to adapt to rapid environmental changes. The history of images helps attention distillation during training, enhancing feature learning. Without this historical context, the model struggles with stable navigation, resulting in suboptimal trajectories.

\begin{table}[t]
\centering
\scriptsize
\begin{tabularx}{\columnwidth}{|c|c|X|X|X|}
\hline
\textbf{Scenario} & \textbf{Method} & \textbf{Success Rate (\%) $\uparrow$} & \textbf{Time to Goal $\downarrow$} & \textbf{Frechet Distance $\downarrow$} \\
\hline
\multirow{6}{*}{Scen. 1} 
& DWA \cite{fox1997dynamic} & 80 & 13.6 & 1.631 \\
& CoNVOI \cite{10802716} & 80 & 21.4 & 0.537 \\
& VANP \cite{10802451} & 70 & 16.3 & 1.321 \\
& \ours{} w/o image sequence & 80 & 14.1 & 0.643 \\
& \ours{} (Ours) & \textbf{100} & \textbf{12.2} & \textbf{0.431} \\
\hline
\multirow{6}{*}{Scen. 2} 
& DWA \cite{fox1997dynamic} & 20 & \textbf{14.2} & 1.219 \\
& CoNVOI \cite{10802716} & 60 & 21.4 & 0.755 \\
& VANP \cite{10802451} & 40 & 17.9 & 1.446 \\
& \ours{} w/o image sequence & 60 & 18.3 & 1.256 \\
& \ours{} (Ours) & \textbf{90} & 15.5 & \textbf{0.732} \\
\hline
\multirow{6}{*}{Scen. 3} 
& DWA \cite{fox1997dynamic} & 30 & \textbf{16.6} & 1.521 \\
& CoNVOI \cite{10802716} & 60 & 27.2 & 1.321 \\
& VANP \cite{10802451} & 50 & 18.4 & 2.271 \\
& \ours{} w/o image sequence & 40 & 21.2 & 1.745 \\
& \ours{} (Ours) & \textbf{90} & 17.2 & \textbf{1.231} \\
\hline
\multirow{6}{*}{Scen. 4} 
& DWA \cite{fox1997dynamic} & 70 & \textbf{18.9} & 2.214 \\
& CoNVOI \cite{10802716} & 50 & 32.4 & 1.452 \\
& VANP \cite{10802451} & 40 & 19.2 & 2.821 \\
& \ours{} w/o image sequence & 60 & 27.6 & 2.139 \\
& \ours{} (Ours) & \textbf{80} & 22.4 & \textbf{1.034} \\
\hline
\end{tabularx}
\caption{\small{The table presents the numerical results of various navigation methods across 10 trials. To assess performance, we utilize three navigation metrics. These metrics are computed as averages across both successful and unsuccessful trials (whether the goal was reached or not).}}
\label{tab:tab2}
\vspace{-15pt}
\end{table}

\no \textbf{Inference Rate}: Our method operates at approximately 20Hz on an Intel i9 processor and an Nvidia RTX 2080 GPU, which is efficient for handling dynamic scenes during robot navigation in real-time.

%% file: 7_Conclusion.tex
\section{Conclusion, Limitations \& Future work}

We introduced \ours{}, a novel Vision-Language Attention Distillation method that enables socially compliant and real-time robotic navigation by distilling knowledge from large Vision-Language Models (VLMs) into a lightweight transformer-based model. By leveraging attention map-level distillation, \ours{} effectively integrates social navigation reasoning from both a pre-trained vision-action model and a VLM, which ensures efficient and adaptive motion planning. Our approach demonstrates significant improvements over SOTA methods in real-world experiments on a Husky robot, with higher success rates and trajectories that align closely with human teleoperated actions. However, as the distilled social priors are learned offline, our attention-guided planner may require further tuning in highly crowded or previously unseen settings. Future work will explore pre-trained models with additional modalities, such as depth and LiDAR, to enhance spatial awareness and robustness. We also plan to investigate online adaptation strategies for real-time refinement, and extensions to long-horizon navigation in complex environments.